\documentclass[letterpaper, 10 pt, conference]{ieeeconf}  % Comment this line out if you need a4paper

\IEEEoverridecommandlockouts                              % This command is only needed if 
\usepackage{amsmath}
\usepackage{graphicx}

\usepackage{array} %table tools
\usepackage{booktabs}
\usepackage{algpseudocode}
\usepackage{cite}

\begin{document}

\title{\LARGE \bf Proximity and Visuotactile Point Cloud Fusion for Contact Patches in Extreme Deformation}

\author{Jessica Yin$^{1*}$, Paarth Shah$^{2}$, Naveen Kuppuswamy$^{2}$, Andrew Beaulieu$^{2}$,\\ Avinash Uttamchandani$^{2}$, Alejandro Castro$^{2}$, James Pikul$^{1*}$, and Russ Tedrake$^{2*}$% <-this % stops a space}
\thanks{$^{1}$Department of Mechanical Engineering and Applied Mechanics and GRASP Lab at University of Pennsylvania, Philadelphia, PA, USA.}%
\thanks{$^{2}$Toyota Research Institute, Cambridge, MA, USA.}% 
\thanks{*Corresponding authors: jessyin@seas.upenn.edu, pikul@seas.upenn.edu, russ.tedrake@tri.global}
\thanks{This work was supported in part by the National Science Foundation Graduate Research Fellowship Program under Grant No. 202095381 and by the National Science
Foundation Emerging Frontiers in Research and Innovation
(EFRI) award \#1935294.}
}

\maketitle

\begin{abstract}
%Equipping robots with the sense of touch is critical to emulating the capabilities of humans in real world manipulation tasks. 
% Visuotactile sensors are a popular tactile sensing strategy due to data output compatible with computer vision algorithms and accurate, high resolution estimates of local object geometry. 
Visuotactile sensors are a popular tactile sensing strategy due to high-fidelity estimates of local object geometry.
However, existing algorithms for processing raw sensor inputs to useful intermediate signals such as contact patches struggle in high-deformation regimes. 
This is due to physical constraints imposed by sensor hardware and small-deformation assumptions used by mechanics-based models.
In this work, we propose a fusion algorithm for proximity and visuotactile point clouds for contact patch segmentation, \textit{entirely independent from membrane mechanics}. 
This algorithm exploits the synchronous, high spatial resolution proximity and visuotactile modalities enabled by an extremely deformable, \textit{selectively transmissive} soft membrane, which uses visible light for visuotactile sensing and infrared light for proximity depth. 
We evaluate our contact patch algorithm in low (10\%), medium (60\%), and high (100\%+) strain states. 
We compare our method against three baselines: proximity-only, tactile-only, and a first principles mechanics model.
Our approach outperforms all baselines with an average RMSE under 2.8 mm of the contact patch geometry across all strain ranges. 
We demonstrate our contact patch algorithm in four applications: varied stiffness membranes, torque and shear-induced wrinkling, closed loop control, and pose estimation. 
%By enabling visuotactile sensing functionality across an unprecedented range of deformation...
% face fundamental challenges with sensing mechanisms that lose functionality beyond small deformations and using soft, hyperelastic materials that are difficult to model accurately in real time mechanics that prevent functionality in higher ranges of deformation, hindering informative contact with cm-scale objects frequently encountered in the real world.
\end{abstract}
\vspace{-1em}
\section{Introduction}

Humans are capable of an exceptionally diverse range of manipulation tasks, from precisely threading a needle to carrying another person.  
Tactile sensing has been studied with the goal of reproducing human-like manipulation abilities in robots.
% motivating the rigorous development of electronic skins and tactile sensors \cite{wang2019tactile, shimonomura2019tactile, shih2020electronic}. 
% Although there has been much progress, achieving the scalability, coverage, resolution, and compliance of human skin in tactile sensors for robots remains a challenge.
A popular tactile sensor paradigm are visuotactile sensors, which utilize high-resolution cameras to leverage advancements in computer vision techniques \cite{yuan2017gelsight, padmanabha2020omnitact, do2022densetact, lepora2021soft, alspach2019soft, lambeta2020digit}.
% Recently, visuotactile sensors  have gained popularity due to ease of fabrication, high-resolution geometry sensing, and compatibility with powerful computer vision techniques. 
% as a tactile sensing strategy due to sensor coverage and resolution that scales with camera field-of-view and pixel count rather than number of sensing elements, as well as compatibility with powerful data processing techniques from computer vision. 
% These sensors use a camera to observe a pattern on a deformable surface, which distorts upon contact. 
% The images of the pattern distortion can inform contact information, such as shear, force, and object geometry. 
\begin{figure}[ht!]

    \centering
    \includegraphics[width=\linewidth]{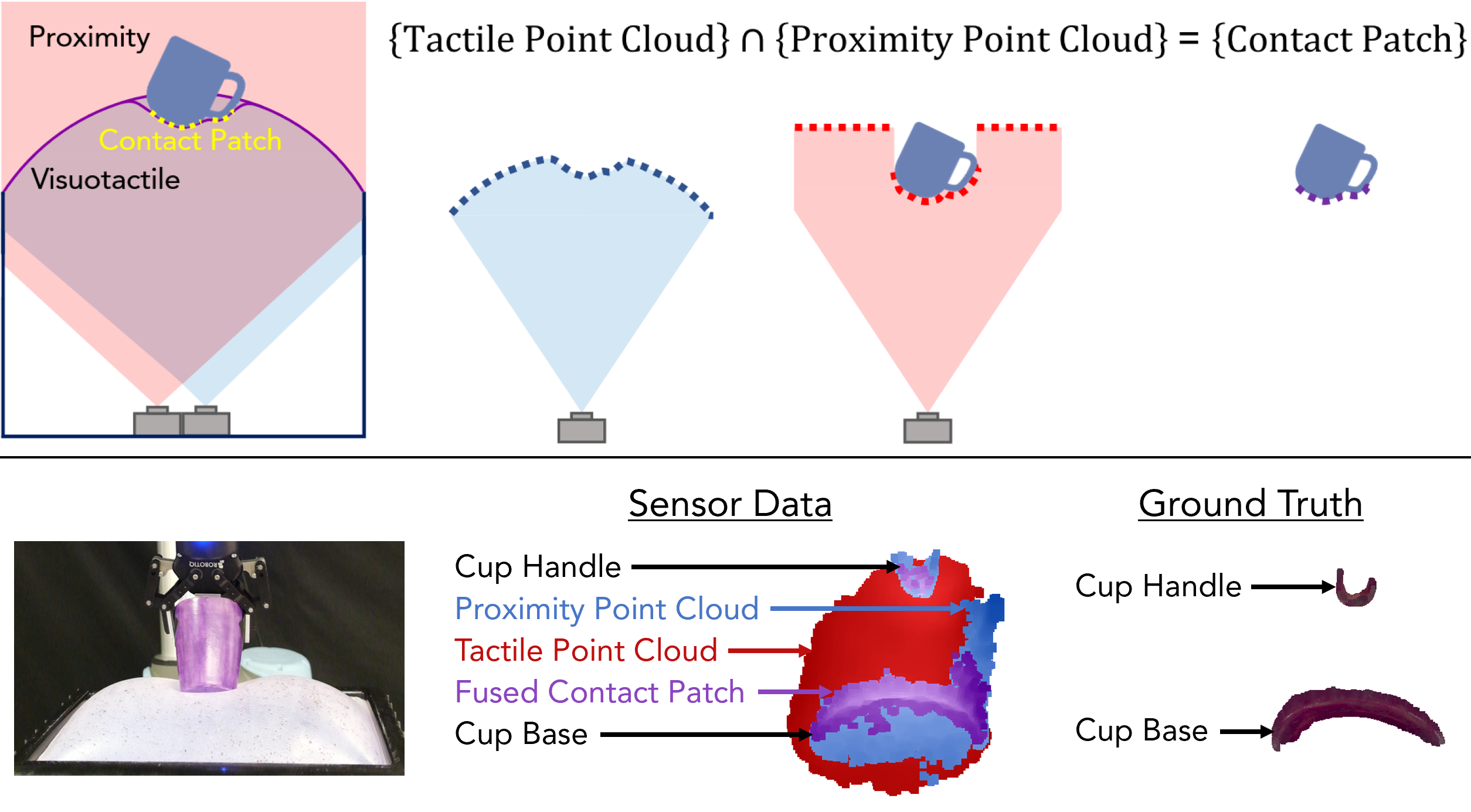}
    \vspace{-2em}
    \caption{
    We propose a mechanics-independent algorithm for estimating contact patches in challenging deformation regimes. We leverage a selectively transmissive soft membrane to provide simultaneous tactile and proximity point clouds. By fusing these two modalities through simply computing the intersection between the two point clouds, we obtain a high quality estimate of the contact patch. Because of our mechanics-independent approach, we avoid the non-linearities that many existing contact patches algorithm struggle with.}
    \label{fig:fig1}
    \vspace{-1.5em}
\end{figure}
However, despite leveraging a soft surface as the contact interface, most visuotactile sensors rely on data processing techniques that lose functionality beyond small deformations, typically less than 1 mm. 
For example, dot tracking and optical flow algorithms used to estimate depth or track motion \cite{du20223d, zhang2018fingervision, li2022continuous, li2023marker} in these sensors are not robust to either occlusions or the large marker displacements that occur during high deformation.
% Additionally, photometric stereo and frustrated-internal-reflection methods \cite{taylor2022gelslim, tippur2023gelsight360} to estimate depth depend on strict lighting conditions that fail when the soft material is highly deformed and too much light is released or the contact area is not evenly illuminated.
% Because these depth estimation methods limit the depth of object penetration, these sensors use a rigid backplate to prevent further deformation. 
This fundamentally restricts how much conformal contact can be made with the object. 

The lack of conformal contact notably leads to small contact patches that are not innately informative for cm-scale objects without additional nontrivial implementations of tactile exploration policies \cite{zhang2023high, suresh2022shapemap}, data-driven pose estimation frameworks \cite{bauza2022tac2pose, suresh2023midastouch}, or integrating external vision sensors \cite{gao2023hand, izatt2017tracking}. 
While small contact patches may be acceptable for small, mm-scale, distinctly featured objects like screws and nuts, general real-world manipulation demands greater versatility and frequent interactions with larger objects. Limited conformal contact can also increase the difficulty of controlling the object.
More compliance can be beneficial as it increases stiction and provides a larger margin of error for robots \cite{oller2023manipulation}. However, existing algorithms are poorly equipped to operate in the these regimes of large compliance.
% Objects are more likely to slip, as interactions with the object closely resemble rigid body interactions .
% In contrast, more compliance increases stiction with the target object and provides a larger margin of error for robots.

% Although larger contact patches are helpful, they introduce new challenges. 
Because deformation occurs beyond the contact patch, it is difficult to detect which parts of the membrane are in contact with the object exclusively from sensing the membrane's deformation. 
The soft materials used to fabricate these compliant sensing surfaces have highly nonlinear mechanical properties that are difficult to model accurately and compute in real time for larger deformations.
% One solution could be to develop a first-principles and linear model for the mechanics of the soft membrane to calculate the contact patch with the object \cite{kuppuswamy2019fast}, but this approach still limits the range of functionality to small deformations. 

In this work, we propose a contact patch segmentation algorithm that fuses proximity point clouds and tactile point clouds to detect contact patches of objects in extreme deformations of the sensing membrane. 
This algorithm leverages the multimodality of a large-scale proximity and visuotactile sensor enabled by a \textit{selectively transmissive soft membrane}.
The proximity depth modality is used to enhance tactile sensing functionality by providing essential object localization and geometry data during contact, independent from the mechanical behavior of the soft membrane. 
With this key information and collocated proximity and tactile depth sensors, the intersection of the proximity and tactile point clouds provides the contact patch (Fig. \ref{fig:fig1}). 
We make two key contributions:
\begin{enumerate}
    \item We propose a novel contact patch segmentation algorithm that is \textbf{\textit{independent from mechanics of the deformable membrane}}. This simple algorithm enables robustness to challenging deformation states, while still being fast enough for real-time control feedback. To the best of our knowledge, this is the first contact patch segmentation algorithm that fuses a point cloud proximity modality with tactile sensing. 
    % and evaluate our contact patch algorithm across low, medium, and high strain membrane states and against three baselines: proximity-only, tactile-only, and a membrane mechanics model.
    % \item We introduce design and fabrication techniques for a multimodal proximity and visuotactile sensor enabled by a selectively transmissive soft membrane.
    \item We demonstrate applications of our contact patch algorithm in four use cases: varied stiffness membranes, torque and shear-induced out-of-plane membrane deformation, closed loop control for whole body manipulation, and pose estimation. 
\end{enumerate}
\vspace{-1em}
% Soft tactile sensors often leverage deformable substrates or sensing membranes with highly nonlinear mechanical properties.
% The stress to strain response of these hyperelastic materials are intrinsically linked to the sensor output in most tactile sensing strategies, such as 
% which make it difficult to preserve sensing functionality throughout the entire range of deformation. 
% Force transduction mechanisms, such as piezoresistive or capactive
\section{Related Work}
Contact patch estimation is often the first step in processing tactile data for downstream tasks such as estimating object pose, contact force distributions and direction, and object classification. 
The three main approaches to contact patch estimation for soft tactile sensors with hyperelastic materials are to apply a threshold based on deformation, use data-driven techniques to classify contact patches, or solve a finite element method (FEM) model online. 

Applying a threshold to membrane displacement is the simplest and fastest method to estimate contact patches \cite{oller2023manipulation}.
A membrane displacement threshold number is manually chosen, and points over the threshold are considered as the contact patch.
However, thresholds require manual tuning specific to the expected deformation and contact interface material, and do not generalize well across a large range of deformations. 

Machine learning methods have been used to predict the contact patch, but they still have significant challenges: (1) the ability to generalize to a diverse set of object geometries, and (2) the laborious process of collecting real-world labelled datasets. 
Because the displacement to force mapping is dependent on object geometry and membrane mechanics, the datasets have to be very  extensive to generalize to all possible objects that could be encountered by the sensor. 
Additionally, the ground truth for each image would have to be generated by a corresponding FEM simulation \cite{narang2020interpreting, sferrazza2019ground}, for which accuracy scales with computation cost.
%This fundamentally limits the accuracy of the model to that of the FEM simulation, which is problematic since FEM simulations have difficulty modelling the complex deformations of a hyperelastic material. 

% Machine learning methods have been proposed to predict the contact patch, but because the contact patch is so dependent on object geometry, it is difficult for the models to generalize beyond the training set of geometry primitives.
% Additionally, with a supervised learning approach [CITE], the ground truth label is the FEM simulation of what the contact patch should look like. 
With the FEM approach, each element of the elastomer is modelled as a hyperelastic material, for which the stress-strain relationship is derived from a strain energy density function \cite{ogden2004fitting}. 
Commonly used hyperelastic models for elastomers include the Neo-Hookean, Ogden, and Mooney-Rivlin formulations. 
% The Neo-Hookean model is the simplest of the three and suitable for low strains, while the Ogden and Mooney-Rivlin models better capture high strains and require more experimental data to fit the model parameters \cite{marckmann2006comparison}. 
However, to determine the stress-stretch curves necessary to use these models, samples of the material must be tested in uniaxial tension, pure shear, and equibiaxial tension \cite{hopf2016analysis}. 
Collecting this data requires specialized equipment and does not generalize well across manufacturing variabilities in different samples.
Most significantly, FEM approaches are computationally expensive and do not run in real-time.
Linear approximations or assumptions of a constant stiffness matrix can be made to speed up runtime \cite{ma2019dense, kuppuswamy2019fast}, but these assumptions do not hold in cases of extreme deformations.
% \begin{itemize}
%     \item kuppuswamy et al
%     \item manipulation with membranes from fazeli
%     \item fem approaches
%     \item learning approaches
% \end{itemize}
% move this to introduction?
\vspace{-0.7em}
\section{Sensor System}
The design and fabrication of the sensor builds upon \cite{yin2022multimodal}. The sensor provides synchronized 640x480 px depth maps in both the proximity and tactile modalities at 30 Hz.

The two internal cameras are the Intel Realsense L515 Time-of-Flight camera (proximity modality) and the Intel Realsense D405 (tactile modality). 
The IR light emitted by the L515 ToF camera travels through the membrane and does not sense the IR-transparent membrane itself. 
Although the membrane is somewhat translucent, the visual texture embedded in the membrane with random dye droplets enables the D405 to produce a point cloud of the membrane. 
The two cameras are placed immediately next to each other at the bottom center of the sensor. 
We design the active sensing region to be located where the FOVs of the D405's stereo depth and L515's ToF camera overlap; the entire exposed membrane is in view of both cameras. 

The soft sensing surface is 355 mm x 205 mm and allows up to 85 mm of vertical displacement from a deflated membrane, before reaching the minimum sensing distances of the internal cameras. The sensor can be pressurized to inflate the membrane; the inflated height directly corresponds to how much the membrane can be vertically displaced. The proximity depth range extends approximately 100 mm beyond the membrane.

\section{Proximity and Visuotactile Fusion for Contact Patch Estimation}
The fusion algorithm outputs a point cloud of the estimated contact patch, which is constructed from the intersection of the tactile and proximity point clouds (Fig. \ref{fig:fig1}).

The data inputs to the algorithm are the proximity camera's RGB image, the proximity depth image, and the tactile depth image. 
The proximity camera's RGB modality, which observes the membrane, is only used to remove noise from the proximity depth data. 
The proximity depth and tactile depth maps are used to compute the contact patch.

\subsubsection{Frame Alignment} First, the proximity RGB and tactile depth images are aligned to the proximity depth map. The proximity RGB and depth images are aligned using the Intel Realsense L515 intrinsics, since the depth and RGB cameras are located on the same device. The proximity depth and tactile depth images are aligned using a manually calibrated homography matrix.
\subsubsection{Proximity Pre-Processing}
Depending on the curvature of the inflated membrane, some areas of the membrane can be infrared reflective and produce noise in the proximity depth data. 
We convert the proximity RGB image to HSV and create a mask that removes any pixels that match the color of the membrane when no objects are in contact or near the membrane surface. 
This mask is tuned to environmental lighting conditions, but applies to all objects.
The mask is then applied to the proximity depth image, which filters out the internal reflections of the emitted infrared light. 
\subsubsection{Contact Patch Estimation} 
The contact patch is estimated by computing the intersection of the tactile and proximity depth images. 
We compute an intersection mask with a pixel-wise comparison of distance values from the tactile and proximity depth maps:
\begin{gather*}
    |d_{p} - d_{t}| \le t|d_{p}|
\end{gather*}
where $d_{p}$ is the proximity distance, $d_{t}$ is the tactile distance, and $t$ is the tolerance.
If the difference in tactile and proximity distance values at a pixel satisfy the inequality, the pixel is considered a match and part of the contact patch.
We manually tune the value of $t$ to produce the contact patch that best matches ground truth, and we found that $t = 0.03$ works well; the tolerance distance values range from 6mm-8mm. 
The intersection mask is applied to the proximity depth image to produce the contact patch estimation. 
We use the proximity depth image instead of the tactile depth image because it is significantly more robust to inconsistent room lighting conditions.
The contact patch depth image is then projected to a point cloud and we apply a statistical outlier rejection filter to remove any extraneous points before outputting the final contact patch point cloud estimation.
Although this step does require some manual tuning of $t$, it generalizes across all strain ranges and objects.

% Then, the tactile depth image is aligned to the proximity depth image.
% A mask of the estimated contact patch is computed with a pixel-wise comparison of distance values from the tactile and proximity depth maps.
% If the difference in distance values at a pixel are within a tolerance of 3\% of the proximity depth value, the pixel is considered part of the contact patch. 
% The mask is applied to the proximity depth image and then projected to a point cloud to produce the contact patch point cloud estimation. 
% We use the proximity depth image over the tactile depth image because the proximity depth data is less susceptible to inconsistent room lighting conditions. 
% Finally, we apply a statistical outlier rejection filter to remove any extraneous points. 

\section{Experimental Methods}
\begin{figure}
    \centering
    \includegraphics[width=\linewidth]{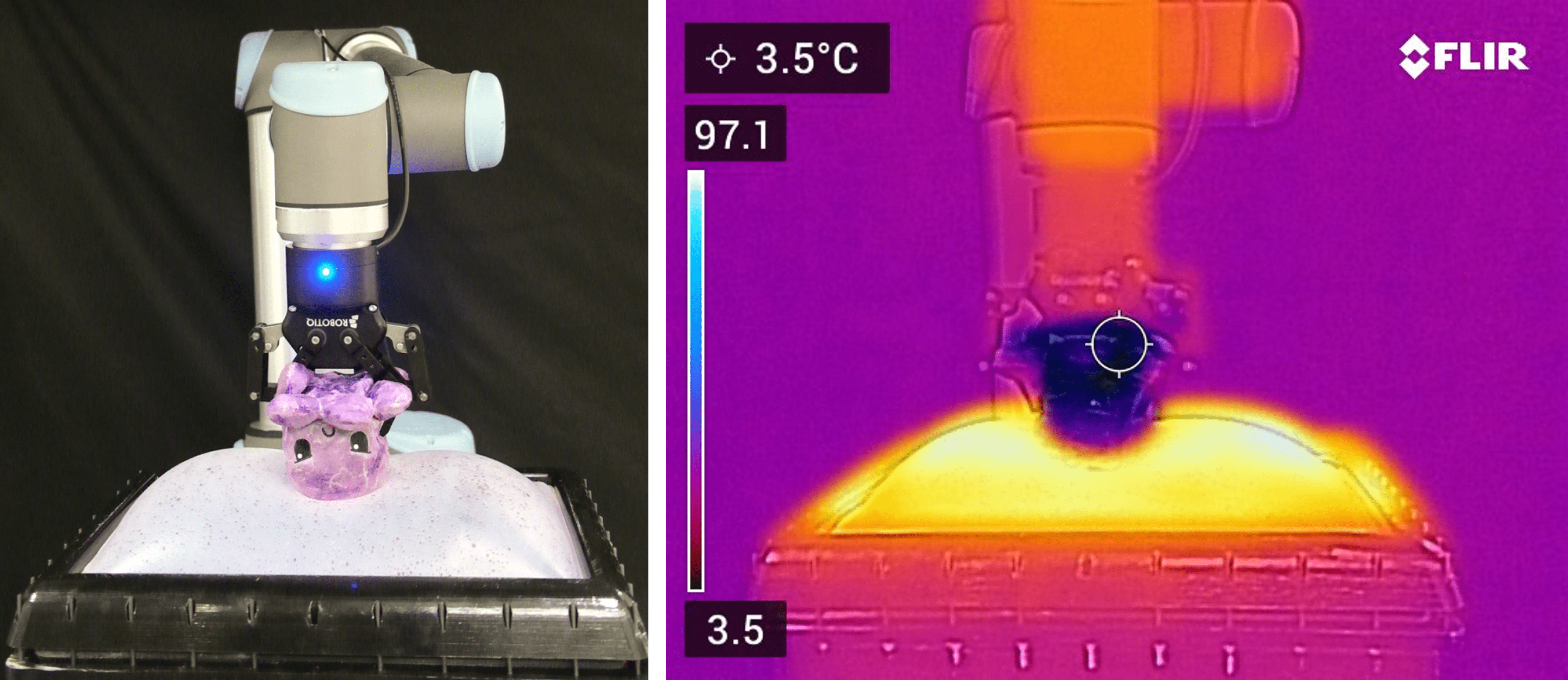}
    \vspace{-2em}
    \caption{The ground truth method for contact patch geometry exploits heat transfer between the object and sensor membrane at the contact interface. The object is painted with thermochromic paint that changes color when it comes into contact with the heated membrane.}
    \vspace{-1.5em}
    \label{fig:gt_fig}
\end{figure}
% \begin{table}
% \caption{Parameters for ground truth method.}
% \label{tab:gt_params}
% \begin{tabular}{|c|c|c|}
% \hline
% \textbf{Object} & \textbf{Membrane Temperature (Celsius)} & \textbf{Contact Time (s)} \\ \hline
% octopus         & 70                                      & 5.5                       \\ \hline
% cup             & 100                                     & 9                         \\ \hline
% cube            & 95                                      & 9                         \\ \hline
% \end{tabular}
% \end{table}
\begin{table}[b!]
    \centering
    \begin{tabular}{lccc}
        \toprule
        Object & Membrane Temperature ($^\circ$C) & Contact Time (s) \\
        \midrule
        octopus & 70 & 5.5 \\
        cup & 100 & 9  \\
        cube & 95 & 9  \\
        \toprule
    \end{tabular}
    \vspace{-1em}
    \caption{\normalfont Parameters for ground truth method.}
    \label{tab:gt_params}
\end{table}

\subsection{Ground Truth}
% It has been challenging to provide an experimental ground truth method for contact patches because the contact patch is always occluded from external viewpoints and surface mounted sensors interfere with the object interaction. 
% Thus, most works use simulation or model-based methods \cite{du2021high, sferrazza2019ground, si2022taxim} to evaluate the contact patch estimation.
% However, our work targets scenarios when simulation or model-based methods fail, so we must use an experimental method for ground truth.

Our ground truth method leverages thermal conduction at the contact interface, similar to \cite{brahmbhatt2019contactdb,lakshmipathy2021contact} (Fig. \ref{fig:gt_fig}).
% Similar methods have been presented to annotate human grasps on objects by capturing heat transfer from human hands with either thermal cameras \cite{brahmbhatt2019contactdb} or thermochromic paint \cite{lakshmipathy2021contact} and RGB-D cameras. 
% However, to the best of our knowledge, a ground truth method has not yet been proposed for the evaluation of contact patch accuracy when simulation or model-based methods fail, such as when hyperelastic materials (i.e., the soft membrane) are in extreme deformation. 
 
% Here, we present the ground truth method used in our work, which leverages thermal conduction at the contact interface \cite{brahmbhatt2019contactdb}.
We paint the object dataset with three coats of thermochromic paint (Elmers), which changes color from purple (cold) to pink (hot) with temperature. 
The objects are cooled to 3.5$^{\circ}$C until the purple color is opaque.
The membrane is then heated to 70$^{\circ}$C-100$^{\circ}$C with a heat gun. 
The precise temperature of the membrane and contact time is tuned per object to prevent noise from thermal radiation; the chosen temperature prevents color change when the object is within 1 mm of the membrane, but enables color change when contact occurs. 
The temperatures and contact time vary per object because each object has a different specific heat capacity (Table \ref{tab:gt_params}).
% which governs the temperature at which the thermochromic paint changes color . 
We monitor the spatial uniformity of the membrane and object temperature with a FLIR thermal camera.  

The thermochromic paint changes color where contact occurs. We scan the object with a high-resolution 3D scanner (Space Spider, Artec) and threshold the scan by color to produce the ground-truth contact patch.  This method captures the maximum contact patch, which we assume to occur when the membrane is at its highest deformation during its interaction with the object.  

% We make the following two assumptions for this ground truth method: 
% \begin{enumerate}
%     \item Because all of the objects are relatively more rigid than the soft membrane, we assume that the object does not deform when contact occurs, so the scan of the object geometry is equivalent to the contact patch geometry. 
%     \item This method captures the maximum contact patch, which we assume to occur when the membrane is at its highest deformation during its interaction with the object.  
% \end{enumerate}

\begin{figure}[b!]
    \centering
    \includegraphics[width=\linewidth]{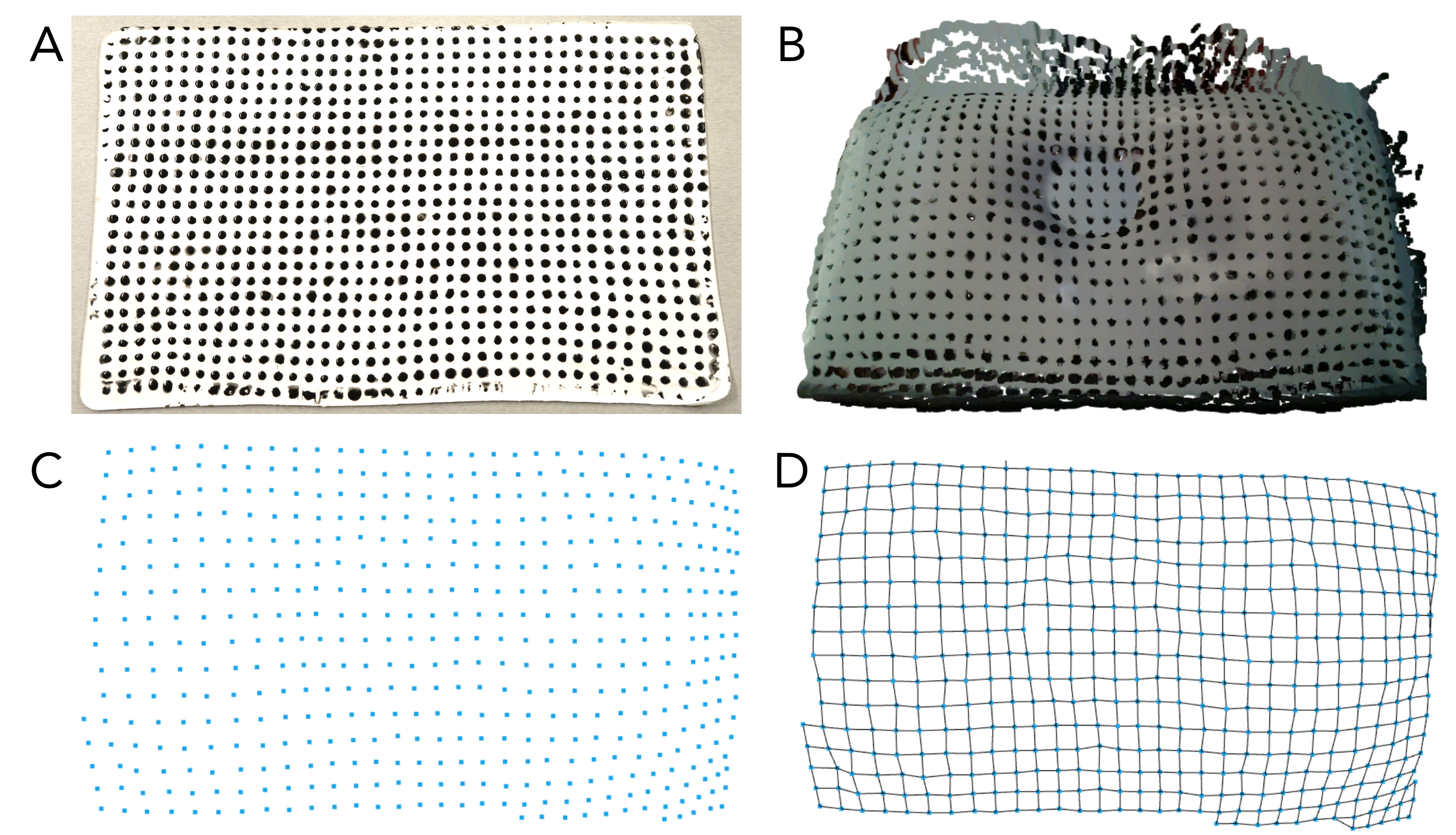}
    \caption{Method used to calculate membrane strain. A) Uniform grid of dots with known spacing is patterned on the membrane. B) Object (octopus, not shown) is pressed into the membrane and the RGB-D point cloud is captured. C) The centroids of the dots from the RGB-D point cloud are isolated. D) The distances between the centroids are measured to calculate strain.  }
    \label{fig:membrane_strain}
\end{figure}
\vspace{-0.5em}
\subsection{Membrane Strain Measurement}
We use the following method to correlate displacement of the object into the membrane to 2D strain in the membrane (Figure \ref{fig:membrane_strain}). 
First, we use a laser-cut stencil to pattern an IR and visibly opaque membrane with a uniform grid of dots with known spacing (10mm).
This membrane is fabricated to have the same thickness and physical dimensions as our sensing membrane.
We use the Realsense L515 to capture an RGB-D point cloud of the patterned membrane as the robot arm applies a vertical displacement with an object from our object dataset.
We measure the distances between the centroids of each dot, which we use to calculate the maximum in-plane strain along the length and width of the membrane. 
The calculated maximum strain is used to find the vertical displacement required to impart the desired strain for our experiments: low (10\%), medium (60\%), high (100\%). 

\subsection{Baselines}

% \begin{figure}[htbp!]
%     \centering
%     \includegraphics[width=0.6\linewidth]{figures/opaque_membrane.png}
%     \caption{This membrane is opaque to IR and visible light. It is used for the tactile-only and membrane mechanics model baselines.}
%     \label{fig:opaque_membrane}
% \end{figure}
For the tactile-only and membrane mechanics model baselines, we use an unpatterned IR-opaque silicone membrane with the same physical dimensions as the selectively transmissive membrane. 
We use the Realsense L515 ToF depth camera to collect data for both of these baselines, since using a ToF depth camera with this sensor system can replicate the Soft Bubble sensor \cite{alspach2019soft} 
and fulfill the input requirement of ToF depth data for the membrane mechanics model. 
For the proximity-only baseline, we use the selectively transmissive membrane and depth data from the Realsense L515 ToF camera.
The computation times for each approach are in Table \ref{table:comp_time}. 
\subsubsection{Tactile-Only}
The tactile-only approach exclusively uses a tactile point cloud and a thresholding scheme to estimate the contact patch. 
In this work, the threshold is implemented as follows: the contact patch is identified as the highest 60\% of deformed points on the membrane, compared to a reference point cloud of an inflated membrane with no object-induced deformation. 
We found that an absolute threshold approach (i.e., points below a certain distance), although common for less deformable visuotactile sensors, gives much worse contact patch estimations and would not be a realistic baseline.  
\subsubsection{Proximity-Only}
The proximity-only approach exclusively uses the proximity point cloud. 
This baseline uses a reference point cloud of the inflated membrane with no object-induced deformation. 
The contact patch is estimated as all points closer to the camera than the reference point cloud. 
\subsubsection{Membrane Mechanics Model}
We use the first principles continuum mechanics model presented in \cite{kuppuswamy2019fast} as a baseline. The model predicts the linear elastic deformation of a mesh of the membrane given the pose and geometry of a rigid object in contact with the sensor. Then, the model is used to solve the inverse problem of estimating the contact patch based on tactile depth and internal air pressure data from the sensor, using a sparse convex Quadratic Program formulation for real-time solutions. The model assumes that the contact surface is frictionless, deforms linearly, and does not wrinkle or fold. 
% \begin{itemize}
%     \item summary from Naveen/Alejandro
%     \item make sure to list assumptions that may be violated during these experiments, such as out-of-plane deformation or high contact loads
% \end{itemize}

% \begin{table}[t!]
% \centering
% \caption{Computation time for one frame of a contact patch estimation, using an Intel Core i7-8650U CPU.}
% \label{tab:latency}
% \begin{tabular}{|c|c|}
% \hline
% \textbf{Algorithm}                               & \textbf{Computation Time (s)} \\ \hline
% tactile-only (baseline)                          & 0.03                          \\ \hline
% proximity-only (baseline)                        & 0.03                          \\ \hline
% fusion (ours)                                    & 0.07                          \\ \hline
% mechanics model (baseline) & 0.3                           \\ \hline

% \end{tabular}
% \label{table:comp_time}
% \end{table}
% \vspace{-1em}
\begin{table}[t!]
    \centering
    \begin{tabular}{lcc}
        \toprule
        Algorithm & Computation Time (s)\\
        \midrule
        Fusion (Ours) & 0.07  \\
        Proximity-Only & 0.03  \\
        Tactile-Only & 0.03\\
        Mechanics Model & 0.3 \\
        \toprule
    \end{tabular}
    \vspace{-1em}
    \caption{\normalfont Time to compute one frame using an Intel Core i7-8650U CPU.}
    \label{table:comp_time}
    \vspace{-4em}
\end{table}
\vspace{-1em}
\section{Evaluation}
\subsection{Experiment Setup}
We measure the accuracy of the estimated contact patch when the membrane is in low (approx. 10\%), medium (approx. 60\%), and high (approx. 100\%+) strain states. 
We use a dataset of three objects: a stuffed octopus, a Rubiks cube, and a cup. 
These objects were chosen to represent a diverse set of geometries: edges (Rubiks cube), curves (octopus), and non-convex, separate contacts (cup). 
Each object is held by a Robotiq gripper mounted to a UR10 robot arm and pushed into the sensor. 
The vertical displacement of the object in each trial induces the desired strain in the membrane: 4 mm for low strain, 24 mm for medium strain, and 120 mm for high strain. For high strain experiments, the object is pressed off-center to allow the uncontacted region to expand outwards and produce high strains in the membrane.

We compare our results to ground truth and three baselines: tactile-only, proximity-only, and the mechanics model-based method. 
The metric for geometric accuracy of the contact patch point clouds is the \textit{average symmetric root mean squared error} (RMSE) of the estimated contact patch point cloud and the ground truth point cloud after manual alignment.
% Although RMSE alone is non-symmetric, we take the average of both RMSE compu
For the membrane mechanics model-based method, we crop the estimated contact patch to exclude the outer edges of the sensor interface since the model cannot consistently predict the ``bulging" behavior of the membrane, leading to spurious estimated contacts.

\begin{figure}[t!]
    \centering
    \includegraphics[width=0.9\columnwidth]{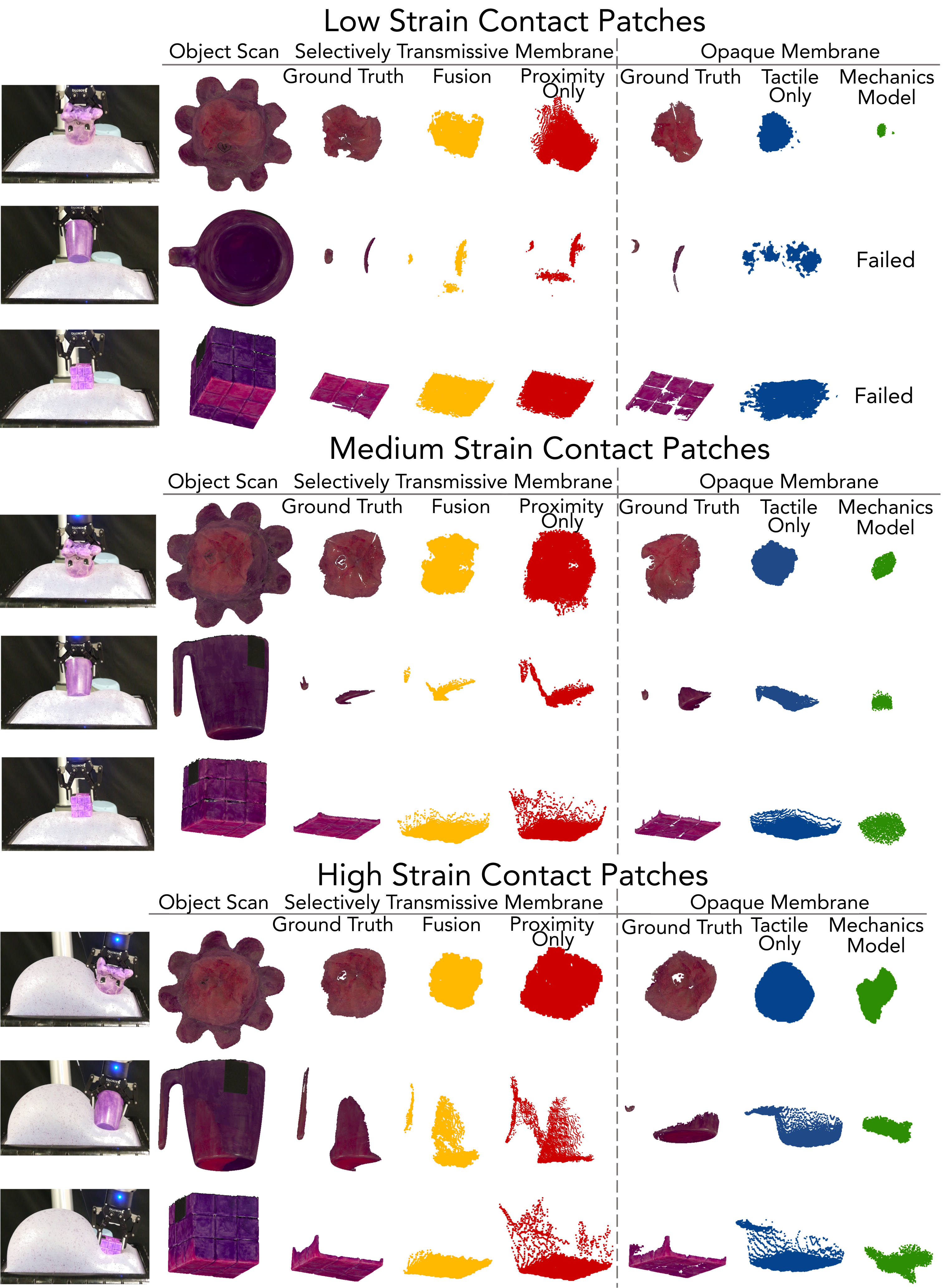}
    \vspace{-1em}
    \caption{Visualizations of the estimated contact patch across low, medium, and high strain states of the membrane. Our proposed algorithm, Fusion, fuses the visuotactile and proximity modalities of the sensor enabled by the selectively transmissive membrane. The baselines for comparison are Proximity Only, Tactile Only, and Mechanics Model. The ground truth is a color-segmented 3D scan of the object, where pink designates contact.}
    \vspace{-2em}
    \label{fig:all_patches}
\end{figure}
% \begin{figure}[htbp!]
%     \centering
%     \includegraphics[width=\linewidth]{figures/all_rmse.png}
%     \caption{RMSE for the contact patch point clouds estimated by the fusion (ours), proximity only, tactile only, and membrane mechanics model algorithms per object. Lower RMSE is better.}
%     \label{fig:all_rmse}
% \end{figure}
% \begin{figure}
%     \centering
%     \includegraphics[width=\linewidth]{figures/avg_rmse_plot.png}
%     \caption{Average RMSE for the contact patch point clouds estimated by the fusion (ours), proximity only, tactile only, and membrane mechanics model algorithms per strain state. Lower RMSE is better.}
%     \label{fig:avg_rmse_plot}
% \end{figure}
\begin{table}[h]
    \centering
    \begin{tabular}{lcccc}
        \toprule
        \textbf{Low Strain (approx. 10\%)} & \multicolumn{3}{c}{Contact Patch RMSE (mm) $\downarrow$} \\
        \midrule
        Algorithm &  octopus & cup & cube\\
        \midrule
        Fusion & \textbf{1.79 }& \textbf{4.96} & 1.34 \\
        Proximity-Only & 2.51 & 7.41 & \textbf{1.27} \\
        Tactile-Only & 3.96 & 6.74 & 1.58 \\
        Mechanics Model & 13.73 & Fail & Fail \\
        \toprule
        \textbf{Medium Strain (approx. 60\%)} & \multicolumn{3}{c}{Contact Patch RMSE (mm) $\downarrow$} \\
        \midrule
        Algorithm &  octopus & cup & cube\\
        \midrule
        Fusion & \textbf{2.64} & \textbf{3.1} & \textbf{1.25} \\
        Proximity-Only & 4.21 & 8.35 & 3.43 \\
        Tactile-Only & 4.32 & 6.07 & 1.75 \\
        Mechanics Model & 9.95 & 10.38 & 6.96 \\
        \toprule
        \textbf{High Strain (approx. 100\%)} & \multicolumn{3}{c}{Contact Patch RMSE (mm) $\downarrow$} \\
        \midrule
        Algorithm &  octopus & cup & cube\\
        \midrule
        Fusion & 3.93 & \textbf{2.85} & \textbf{1.57} \\
        Proximity-Only & \textbf{2.08} & 6.63 & 4.28 \\
        Tactile-Only & 4.1 & 4.2 & 3.79 \\
        Mechanics Model & 8.31 & 5.9 & 8.86 \\
        \toprule
        \textbf{Overall Average for Strain Range} & \multicolumn{3}{c}{Contact Patch RMSE (mm) $\downarrow$} \\
        \midrule
        Algorithm &  low & medium & high\\
        \midrule
        \textbf{Fusion} & \textbf{2.7} & \textbf{2.33} & \textbf{2.78} \\
        Proximity-Only & 3.73 & 5.33 & 4.33 \\
        Tactile-Only & 4.09 & 4.05 & 4.03 \\
        Mechanics Model & 13.73 & 9.1 & 7.69 \\
        \toprule
    \end{tabular}
    \vspace{-1em}
    \caption{\normalfont Results from experiments across low, medium, and high strain ranges. Average symmetric RMSE is used to calculate error.}
    \label{table:results}
    \vspace{-4em}
\end{table}
\vspace{-0.5em}
\subsection{Results and Analysis}
% \subsection{Low Strain Experiments}
% We induce low strain in the membrane by commanding the UR10 robot arm to press the object 4 mm into the membrane. 
% The initial maximum height of the membrane is 314 mm from the internal cameras at the base of the sensor. 
The contact patch estimations are shown in Figure \ref{fig:all_patches} for a qualitative evaluation of contact patch geometry. The average symmetric RMSE analysis per strain state is shown in Table \ref{table:results}.
Although our proposed fusion algorithm is simple, it generalizes well across our object dataset and strain range, outperforming all of the baselines on average (Figure 7). 
Fusing both the tactile and proximity modalities is key to overcoming the shortcomings of each modality: the proximity modality tends to overestimate contact while tactile thresholding tends to underestimate contact. 

We highlight the fusion algorithm performance for the cup, which requires segmenting two separate contacts with the base and the handle. For the medium and high strain experiments, the fusion algorithm is the only method capable of accurately distinguishing theses separate contacts. 

Generally, the proximity modality overestimates contact because it cannot directly sense contact with the object. Tactile thresholding underestimates contact because there is not one threshold that can be applied across all strain ranges. Additionally, tactile thresholding struggles to distinguish two separate contacts. 

There are only two specific cases where a baseline has a lower RMSE than the fused contact patch: the proximity-only estimations in the high-strain octopus contact patch and the low-strain cube patch. 
With both of these cases, the proximity-only baseline overestimation of the contact patch better captured the ground-truth contact patch geometry than the underestimation from the fusion algorithm.
In all other cases, the fusion algorithm consistently produced the most accurate contact patch with the lowest RMSE error. 

In some low strain experiments, the membrane mechanics model failed to produce an estimate of the contact patch, because the deformation is within the same magnitude as the simulated depth camera noise and thus indistinguishable. We note that the contact patch geometry estimate is dominated by the coarseness of the mesh. 
\begin{figure}[b!]
    \centering
    \includegraphics[width=\columnwidth]{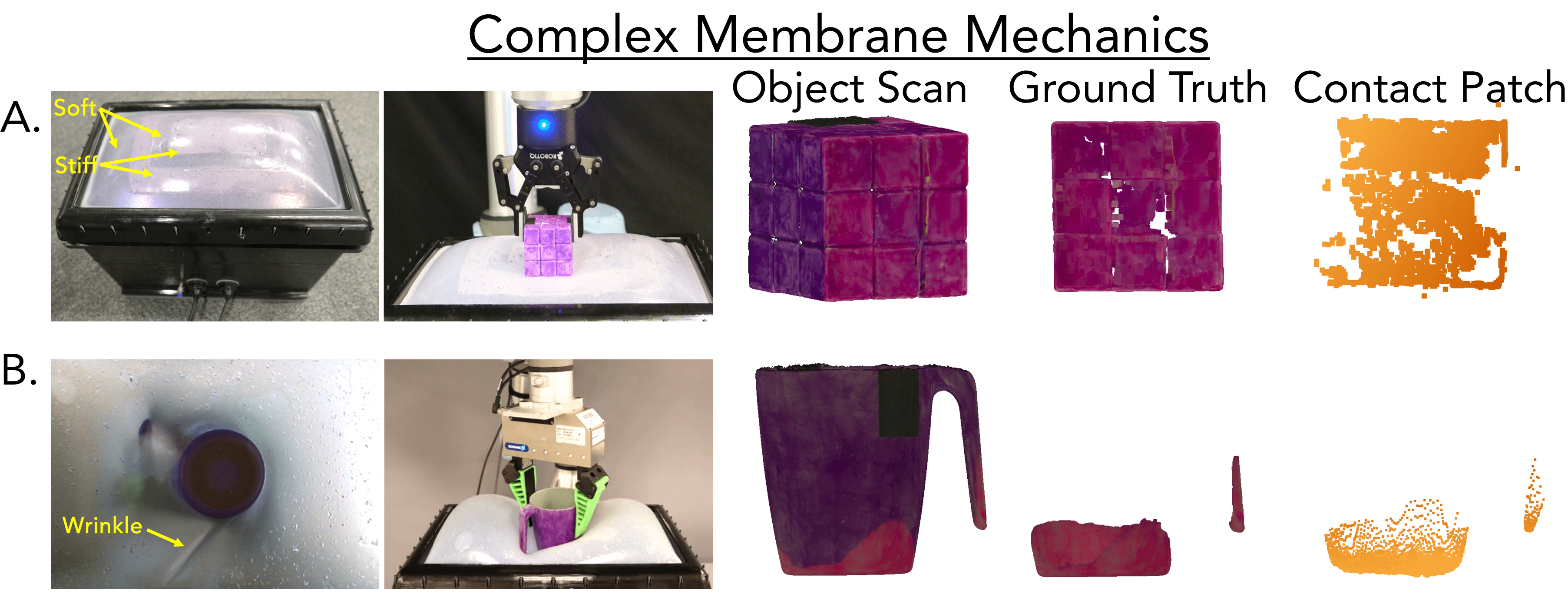}
    \vspace{-2em}
    \caption{We demonstrate the proposed fusion algorithm for contact patch estimation in two applications with complex membrane mechanics: A. a varied stiffness membrane and, B. membrane wrinkles, or out-of-plane deformations. }
    \label{fig:complex_mechanics}
\end{figure}

\section{Demonstrations}
% We demonstrate several applications of the contact patch fusion algorithm to highlight the usefulness of fast and accurate contact patch estimation during complex membrane deformations: varied stiffness membranes, torque and shear-induced membrane wrinkling, closed loop control, and pose estimation.  
\subsection{Complex Membrane Mechanics}
\vspace{-0.5em}
\subsubsection{Varied Stiffness Membranes}
Active and passive variable stiffness materials \cite{pikul2017stretchable, baines2023programming, campbell2022electroadhesive, baines2022multi} can be used for real-time optimization of robot end effector shapes and stiffnesses. Here, we demonstrate the robustness of our sensor's multimodality and contact patch estimation to variations in the sensing membrane's stiffness patterns (Fig. \ref{fig:complex_mechanics}A).
% such as directing the inflation of the membrane to a flatter shape for larger contact areas with less force, rather than a uniform and curved expansion of the membrane. 
% Additionally, actively controlled variable stiffness materials \cite{campbell2022electroadhesive, baines2022multi} are a rich area of soft robotics research and could potentially be used to change the stiffnesses and shapes of robot end effectors to suit different tasks in real time. 
% Because the selectively transmissive membrane exclusively relies on optical properties to enable the tactile and proximity modalities, the stiffness of the membrane can be varied with no impact to the sensing functionality. 
% This sensing approach, used in conjunction with the mechanics-independent proximity and tactile fusion contact patch algorithm, enables contact patch estimation for arbitrary patterns of varied stiffness in the sensing membrane. 
% Because both our sensing mechanism and d
We fabricate a varied stiffness membrane by using two base silicones, Dragonskin 10 Fast and Ecoflex 00-35 Fast (Smooth-On), with different Shore hardnesses (Figure 8). Dragonskin 10 is approximately 8x harder than Ecoflex 00-35. 
The stiff and soft regions are cured together to form one cohesive varied stiffness membrane.
% The two stiff regions are made from a stiffer silicone, Dragonskin 10 Fast (Smooth-On), which has a Shore hardness of 10A. 
% The two soft regions are made from Ecoflex 00-35 Fast (Smooth-On), which has a Shore hardness of 00-35. 
% The same fabrication method presented in Section IIIA is used to produce the varied stiffness membrane, albeit with the above specified base elastomers.

Here, the advantage of our approach is generalization across different stiffness patterns.
The proximity-only and tactile-only baseline methods would require tuning for each membrane pattern due to the variance of inflated shapes. 
The membrane mechanics model would require generating a new mesh that matches both the stiffness pattern and mechanical properties of the different membrane patterns. We run the medium strain cube experiment with the varied stiffness membrane and find the RMSE is 1.09 mm.
% A machine learning approach would require more data for each new membrane pattern. 
\subsubsection{Torque and Shear-Induced Wrinkling}
A highly deformable contact interface that tolerates a large range of shear and torque before slipping is key to maintaining stiction with a target object.
This feature can be critical in mitigating errors in robot manipulators by reducing the requirement for precise perception and control during a task, or simplifying planning by minimizing regrasping or repositioning the object. 
However, these torques and shear forces can create wrinkles in the deformable membrane surface, which are very difficult to model even with computationally expensive FEM approaches.
In this demonstration, we highlight the advantage of our mechanics-independent, fusion contact patch algorithm to estimate the contact patch of a cup while the membrane is wrinkled (Fig. \ref{fig:complex_mechanics}B). 
The robot arm presses the cup into the surface of the sensor and rotates it 45$^{\circ}$, creating wrinkles behind the cup handle and along the sides of the cup.
The estimated contact patch is shown in Figure 8. In comparison with ground truth, the RMSE of the estimated contact patch is 3.49 mm. The geometry of the contact patch qualitatively matches well with ground truth, accurately capturing the side of the cup handle that is in contact with the membrane despite the wrinkles.

\begin{figure}[bp!]
    \centering
    \includegraphics[width=0.7\linewidth]{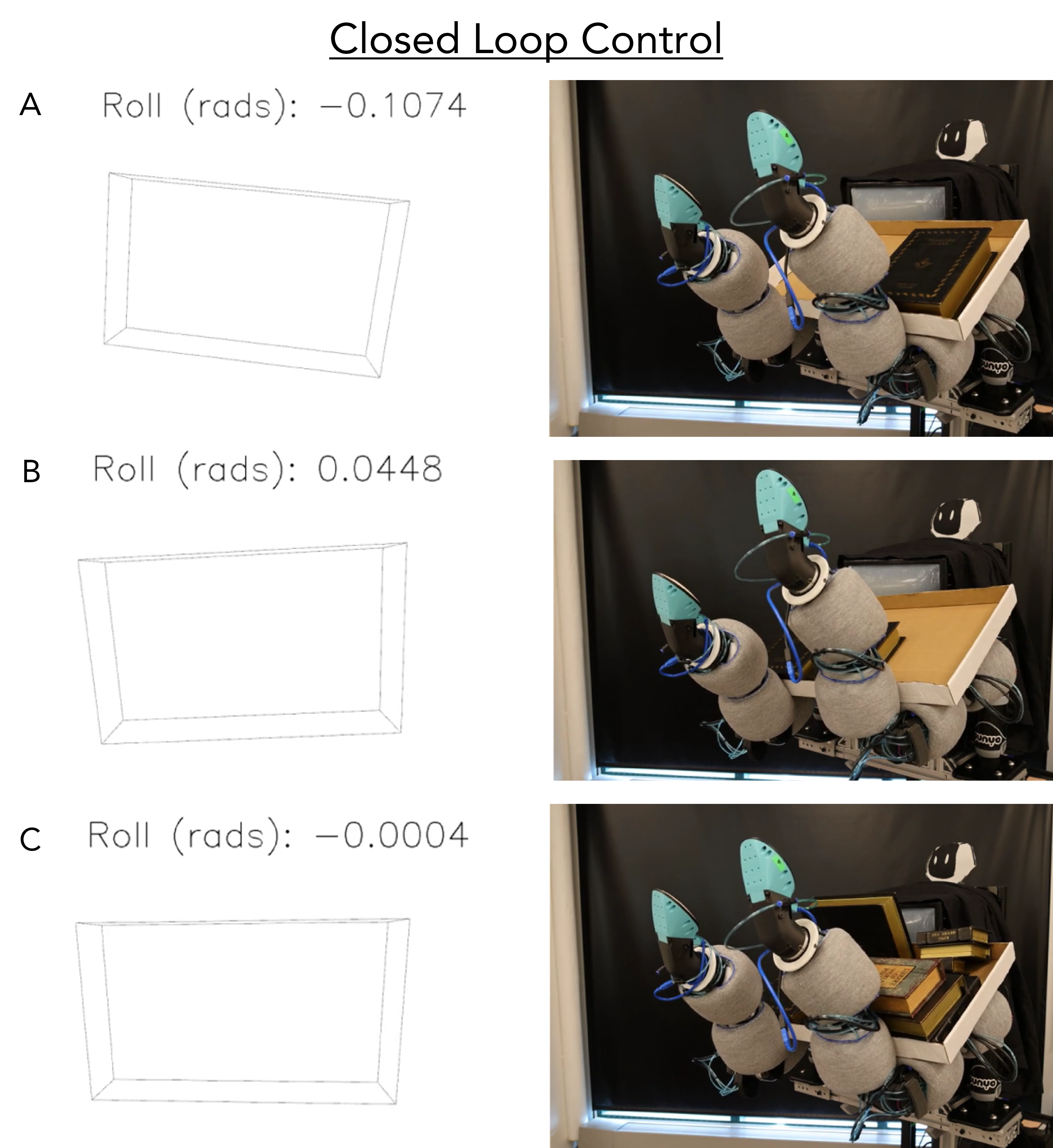}
    \vspace{-1em}
    \caption{The sensor is integrated on a dual arm robot platform as the chest of the robot. The sensor estimates the angle from the contact patch of the tray of books in order to balance the tray as more books are added. This task demonstrates the real-time speed and integration of the proximity and tactile fusion algorithm with a closed-loop controller. A) The tray tilts towards the right with the weight of the book. B) The tray tilts towards the left as the book slides towards the left. C) The robot holds the tray level to balance the stack of books. }
    \label{fig:book_demo}
\end{figure}
\vspace{-0.5em}
\subsection{Closed-Loop Control for Whole-Body Manipulation}
\vspace{-0.3em}
% *need to motivate why large contact patches are useful for whole-body manipulation
We demonstrate the real-time speed and integration of the proximity and tactile fusion algorithm with a closed-loop controller to balance a stack of books (Fig. \ref{fig:book_demo}). 
% and physical robot system for a whole body manipulation task: 
The robot system, Punyo-1 \cite{goncalves2022punyo}, consists of the visuotactile and proximity sensor placed on the robot's torso, a Kinova Jaco robot arm on each side of the sensor with soft padding, and a Puget computer. 
An empty cardboard tray is held by Punyo-1 and pressed into the sensorized torso. 
As each book is placed on the tray held by the robot, the tray's center of mass changes and the robot adjusts the roll angle of the tray to balance the books. 
The angle of the tray of books is sensed with the proximity and visuotactile fusion contact patch algorithm. 

The sensor runs at 30 FPS and the contact patch is calculated for each frame. 
To calculate the angle of the tray, we use principal component analysis (PCA) on the contact patch. 
We found that PCA of the contact patch gives less noisy estimations of the tray position compared to the proximity-only modality, which has added noise from sensing the books. 
The estimated angle of the tray is given to the whole-body PD controller at a frequency of 29 Hz.
We use two arm motion primitives to adjust the tray angle.
% \begin{itemize}
%     \item Summary of controller from Paarth
%     \item mention motion primitives, speed (Hz) of controller?
% \end{itemize}
\vspace{-0.7em}
\subsection{Pose Estimation}
\begin{figure}
    \centering
    \includegraphics[width=0.8\linewidth]{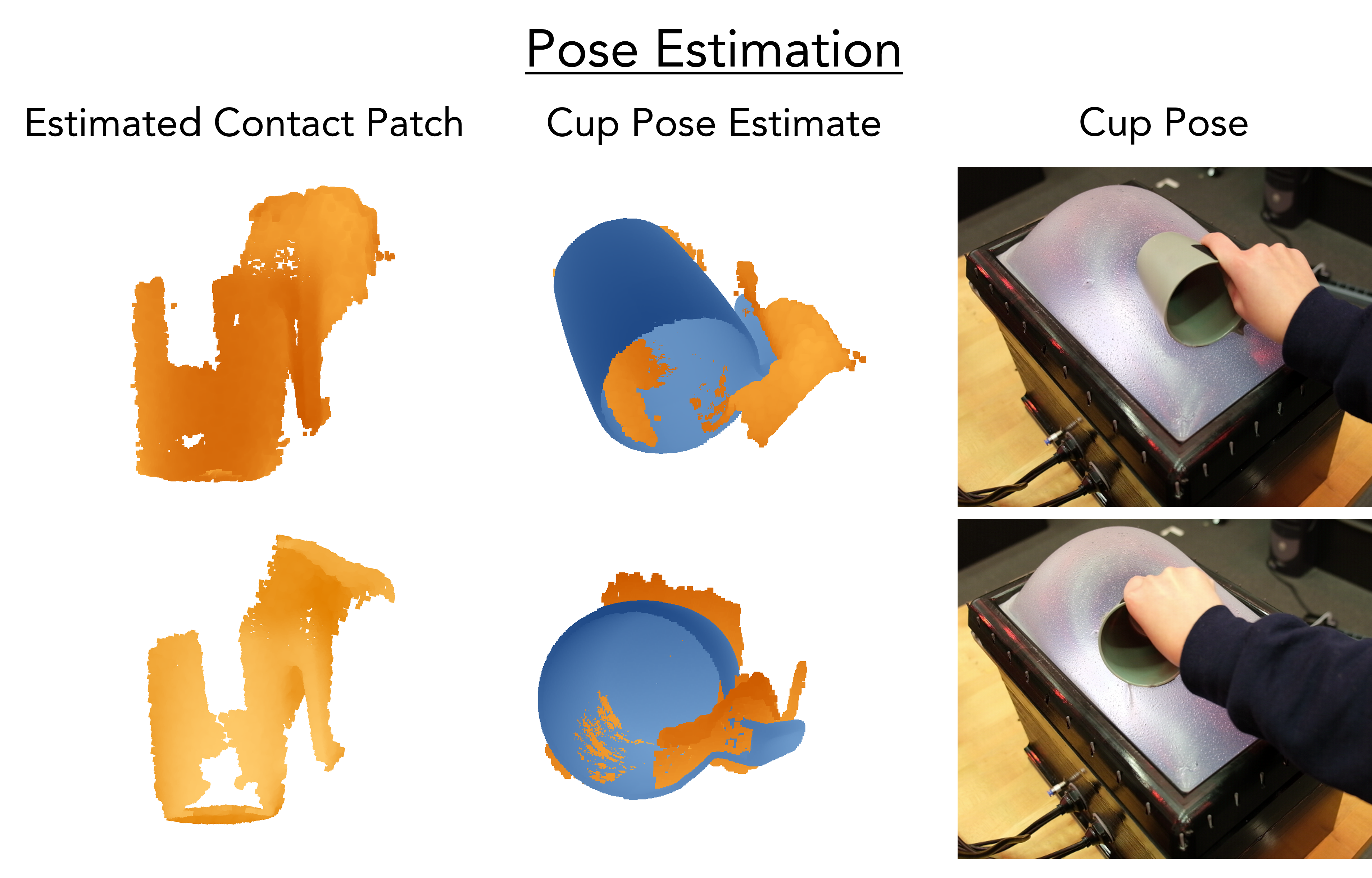}
    \vspace{-1em}
    \caption{The cup simultaneously rotates about the handle and translates across the membrane. The pose estimate from the contact patch qualitatively matches very well. The estimated contact patch has an atypical amount of noise near the top of the cup, which comes from the proximity estimation of the hand, due to adverse lighting conditions.}
    \label{fig:cup_pose}
    \vspace{-1.5em}
\end{figure}
In this demonstration, we show that our contact patches are compatible with a standard pose estimation pipeline (Fig. \ref{fig:cup_pose}).
The greater conformal contact allows for large and dense contact patch point clouds that are easy to use with a standard Iterative Closest Point (ICP) implementation from Open3D \cite{Zhou2018} to estimate pose. 
Exclusively using contact patches for pose estimation, rather than only the proximity modality, could be advantageous in cluttered environments where an additional segmentation step would be necessary to isolate the object of interest.
We estimate the pose of a cup for 5 s as the cup simultaneously rotates about the handle and translates across the sensor surface.
% Because of stiction between the cup and the sensor surface, wrinkles are created in the membrane during the cup trajectory.
We assume a known model of the cup geometry.
In the first frame of the trajectory, the cup is manually registered to the estimated contact patch.
For the following frames that measure the rest of the trajectory, the relative transformation of the cup is estimated for each timestep. The pose estimation tends to become unstable when the cup handle is not observable. Qualitatively, the pose estimate of the cup matches well and is fairly stable across the 5 s of data. A future application of this work could be tactile SLAM for simultaneous object pose estimation and reconstruction \cite{suresh2021tactile, zhao2023fingerslam}.

\vspace{-0.3em}
\section{Conclusion and Future Work}
\vspace{-0.3em}
In this study, we introduced a proximity and visuotactile point cloud fusion algorithm to detect contact patches with soft membranes across an unprecedented range and type of deformation. 
The key idea of the algorithm is that the contact patch can be identified as the set of points where the proximity point cloud and tactile point cloud intersect.
Because the proximity point cloud provides localization and object geometry independent of membrane mechanics, our method is not limited by the challenge of accurately modeling the complex deformations in hyperelastic membrane materials.
% This is not possible for visuotactile sensors that rely exclusively on the tactile modality. 
% The tradeoff between modeling these behaviors for highly deformable sensing surfaces or severely restricting deformation for linear assumptions remains for visuotactile sensors that rely exclusively on the tactile modality.

% We also presented the design and fabrication techniques for a selectively transmissive soft membrane that enables the dual sensing modalities. 
% The formulation of the soft membrane leverages the insolubility of a visibly-colored, IR-transparent aqueous dye solution in a highly elastic silicone elastomer to create visual texture for tactile sensing with a stereo RGB depth camera and enable proximity sensing for an internal ToF depth camera.
% The fundamental sensing mechanism of selective light transmission is robust to manufacturing variances in the membrane, so the proposed sensor fusion algorithm requires no tuning for different batches of membranes. 
% Additionally, the fabrication process uses accessible, off-the-shelf materials and produces a physically resilient and easily repairable sensing interface. 

% Furthermore, we
% We conduct a study of the contact patch geometry accuracy for a diverse object dataset across three regimes of deformation: low strain, medium strain, and high strain. 
We found that our proposed proximity and visuotactile fusion algorithm offered the best performance across all three strain categories. 
We demonstrated the use of this algorithm with three applications: complex membrane mechanics such as varied stiffness and wrinkled membranes, closed loop control for balancing a stack of books with a dual arm manipulation platform, and pose estimation. 
% Moving forward, we would like to explore modifying the sensor design to achieve higher sampling frequencies for more dynamic manipulation tasks. The sensor is limited by the frame rate of the proximity ToF depth camera, to maintain synchronous sampling with the tactile stereo RGB depth camera. The selective transmission sensing mechanism does not solely rely on using a ToF depth camera; the requirement is for an IR camera. A high-speed IR camera (stereo or mono) could be used to replace the ToF depth camera to increase the speed of the entire sensing system. 
% This work could also be extended by using this contact patch algorithm and selective light transmission sensing mechanism for variable stiffness surfaces.

One future direction is to explore a fingertip form factor. The current sensor size is driven by minimum sensing distances of the internal depth cameras, which can be replaced with shorter-range sensors for a significantly smaller overall package.
Another future direction is to investigate using this sensor and algorithm with an actuating surface that changes stiffness and shape in real-time. A shape-changing end effector can be optimized for interactions with different objects. Our algorithm is uniquely suited to mitigate the challenges of using soft materials with variable stiffnesses and shapes.

% Another future direction for this work is to further investigate the use of proximity depth sensing in perception for manipulation. In this paper, we presented proximity as a modality to support and enhance tactile sensing functionality. It would be interesting to study the benefits of proximity sensing as a more independent modality in a manipulation pipeline that incorporates vision and tactile sensing.

\vspace{-0.5em}
\section*{Acknowledgment}
We thank the PrOps team, Aimee Goncalves, Dr. Jose Barrieros, Dr. Aykut Önol, Eric Cousineau, and Alex Alspach at TRI for helpful discussions, assistance with experiments, and hardware design advice. We thank William Yang and Greg Campbell at the Penn GRASP Lab for assistance with experiments and helpful discussions. 

%\section*{References}
\bibliographystyle{IEEEtran}
\bibliography{main}

\end{document}